\documentclass[10pt]{article} 
\newif\ifsubmission
\submissionfalse
\ifsubmission
  \usepackage{tmlr}
\else
  \usepackage[preprint]{tmlr}
\fi

\usepackage{amsmath,amsfonts,bm}

\def\eqref#1{equation~\ref{#1}}

\def\1{\bm{1}}

\DeclareMathAlphabet{\mathsfit}{\encodingdefault}{\sfdefault}{m}{sl}
\SetMathAlphabet{\mathsfit}{bold}{\encodingdefault}{\sfdefault}{bx}{n}

\usepackage{hyperref}
\usepackage{url}
\usepackage{xcolor}
\usepackage{graphicx}
\usepackage{booktabs}
\usepackage{amsmath}
\usepackage{amssymb}
\usepackage{placeins}
\usepackage[capitalize]{cleveref}
\crefname{appendix}{Appendix}{Appendices}
\crefname{section}{Section}{Sections}
\crefname{figure}{Figure}{Figures}
\crefname{table}{Table}{Tables}

\newcommand{\Nninety}{N_{90}}
\usepackage{etoolbox}
\AtBeginEnvironment{table}{\setlength{\belowcaptionskip}{0.5em}}
\definecolor{zerocarry}{HTML}{2A8C82}
\definecolor{poscarry}{HTML}{3E6FA8}
\newlength{\baseboxmin}
\newsavebox{\baseboxcontent}
\newcommand{\basebox}[2]{{\setlength{\baseboxmin}{1.2em}\sbox{\baseboxcontent}{\vphantom{0g}#2}%
  \setlength{\fboxsep}{1.5pt}\colorbox{#1!30}{\ifdim\wd\baseboxcontent<\baseboxmin\makebox[\baseboxmin][c]{\vphantom{0g}#2}\else\vphantom{0g}#2\fi}}}
\newcommand{\zc}[1]{\basebox{zerocarry}{#1}}
\newcommand{\pc}[1]{\basebox{poscarry}{#1}}
\usepackage[most]{tcolorbox}
\definecolor{defnavy}{HTML}{2E3B5C}
\newcommand{\defbox}[2]{\begin{tcolorbox}[enhanced, colback=defnavy!5, colframe=defnavy!55, boxrule=0.6pt, arc=2pt, left=6pt, right=6pt, top=4pt, bottom=4pt, before skip=8pt, after skip=8pt]\textbf{#1}\enspace\itshape #2\end{tcolorbox}}

\title{The Long Delay to Arithmetic Generalization}

\author{\name Laura Gomezjurado Gonzalez \email lpgomez@stanford.edu \\
      \addr Stanford University}

\def\month{MM}  
\def\year{YYYY} 
\def\openreview{\url{https://openreview.net/forum?id=XXXX}} 

\begin{document}

\maketitle

\begin{abstract}
Large language models today still make mistakes on multi-digit arithmetic when they do not have tools, and part of the reason is how numbers are split into tokens. Arithmetic is also a setting where the correct answer to every input and the algorithm that produces it are already known, so training transformers on it lets us study what they learn against a ground truth. These models often fit their training set long before they generalize to held-out inputs, and this delay, known as grokking, decreases as the training set grows. But, how does the representation of numbers change the amount of data a transformer needs to generalize? We study one-step Collatz prediction, in which a transformer reads the digits of an integer in a chosen base and outputs the digits of the next term in its Collatz sequence. We find that the long delay appears only when the training set has few different integers, and that the numeral base determines how many integers are sufficient. In bases where each output digit depends only on a few neighbouring input digits, because the carry between positions depends only on the current digit, a transformer needs several times fewer training integers. We predicted this ordering before training, and input length does not explain it. The same carry structure predicts which bases are easy for a second Collatz map, for integer division, for a decoder-only transformer, and across three model sizes. Finally, a linear probe recovers the low-order residues of the input as accurately from a randomly initialized encoder as from a trained one. Thus, high probe accuracy on such features does not reliably show that a model has learned to compute them. More broadly, we show that the choice of numeral representation affects sample efficiency in ways we can predict from the structure of the task before training.
\end{abstract}

\section{Introduction}
\label{sec:intro}

\begin{figure}[t]
    \centering
    \includegraphics[width=0.92\linewidth]{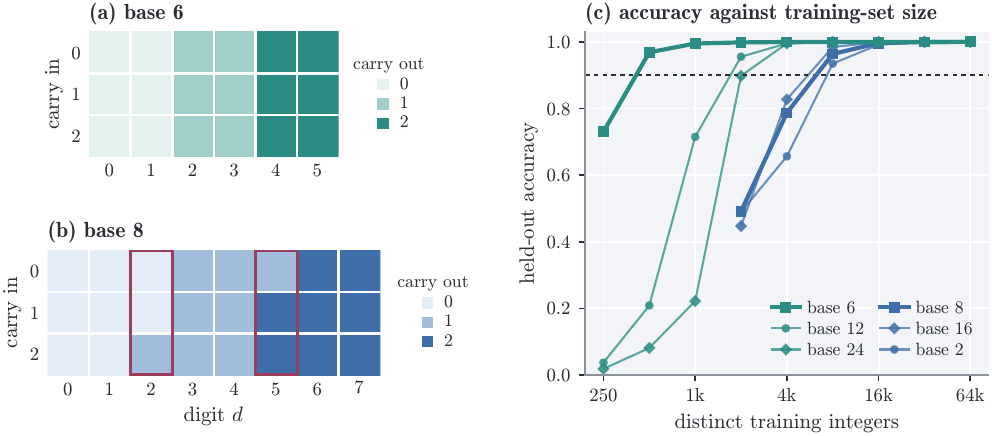}
    \caption{\textbf{In base 6 the carry that $3n+1$ passes on depends only on the current digit, and bases divisible by 3 reach 90\% accuracy with 3 to 18 times fewer training integers than bases 2, 8 and 16.} Panels (a) and (b) show the outgoing carry of $3n+1$ for every digit and incoming carry. In base 6 each column has a single shade, so the outgoing carry is the same for every incoming carry. In base 8 the two outlined columns pass the incoming carry on, which lets a carry travel across several digits. Panel (c) shows the mean over 2 seeds of the best held-out accuracy within 30,000 steps. Teal marks bases divisible by 3 and blue marks the other bases, the thick lines are the bases of panels (a) and (b), and the dashed line marks 0.9.}
    \label{fig:carry}
\end{figure}

When we train transformers on arithmetic tasks (e.g.,~modular addition or modular division), they often fit their training set before they generalize to held-out data points, which is known as grokking. \citet{lr01} reported this pattern and found that the number of optimization steps needed to generalize grows quickly as the fraction of data used for training decreases. \citet{lr05} explain grokking by showing that, once the training set is large enough, a solution that computes the rule reaches low loss with a smaller parameter norm than one that memorizes the training set, and \citet{lrZhu} measure this threshold for language models. In arithmetic tasks, the data a model needs also depends on how numbers are represented, such as their format, digit order, and numeral base. Whether sequence models learn addition depends on the surface form of numbers \citep{lr10}, reversing the output digits lowers the number of examples addition needs \citep{lr11}, and the divisors of the numeral base determine which greatest common divisors a transformer learns \citep{lr25}.

\citet{lr26} trained transformers on hundreds of millions of examples of a long Collatz step and found that accuracy depends strongly on the base. Their models trained on far more examples than they could memorize, so these differences between bases hold when data is abundant. In this paper, we explore how the base changes the number of different training data points a model needs, since that number sets the length of the delay between fitting the training set and generalizing. \citet{lrDawes} found that the structure of the carry operation predicts how well a recurrent network learns addition from a fixed training set, which also led us to explore carries.

We study one step of the Collatz map, in which an encoder-decoder transformer reads the base-$b$ digits of an integer $n$ and writes the digits of $T(n)$, which equals $n/2$ for even $n$ and $3n+1$ for odd $n$ (\cref{sec:setup}). Every model achieves almost perfect accuracy on its training set early in training. When the training set contains few different integers, held-out accuracy then rises slowly for tens of thousands of steps, and when it contains many, held-out accuracy rises at about the same time (\cref{fig:delay}). So the long delay appears only below a threshold of different training integers, which matches the critical dataset size described by \citet{lr05}.

We then train the model in six bases and measure, for each base, how many distinct training integers it needs to reach 90\% held-out accuracy (\cref{fig:carry}). Before training, we computed two quantities for each base, the length of its inputs and the carry measure $\eta_1$, which is zero when the carry that $3n+1$ passes to the next digit depends only on the current digit (\cref{fig:carry_example}). From input length, we predicted that bases with shorter inputs would need fewer integers, and from $\eta_1$, that bases with $\eta_1 = 0$ would. The bases with $\eta_1 = 0$ need several times fewer integers, and this includes base 6, whose inputs are longer than those of base 8. The same carry structure let us predict the easy bases for the map with $5n+1$, for integer division by 3, for a decoder-only transformer and at three model sizes (\cref{sec:base}). Every base achieves full accuracy with enough integers, so the base changes how much data the model needs. In the long Collatz step studied by \citet{lr26}, some bases remain inaccurate with abundant data.

To see what a model learns below the threshold, we froze the encoder of a trained model and trained a new decoder on top of it (\cref{sec:encoder}). With a small training set, the new decoder generalizes within a few thousand steps on the trained encoder and more slowly on a frozen random encoder. With a large training set, it generalizes as fast on the random encoder as on the trained one.

We finally study linear probing, a common interpretability method used to argue that a model encodes a feature, and ask what it shows about a model before it generalizes. A linear probe is a linear classifier fit to the model’s internal states to recover a task feature, and its accuracy on held-out inputs is read as how much of that feature the model encodes. We call this accuracy the \emph{linear decodability} of the feature. The low-order residues of $n$ are determined by the last few input digits, and they are perfectly decodable from a randomly initialized encoder and from the input digits themselves (\cref{sec:decoding}). Only features that combine several digit positions, such as the leading digit of $T(n)$, separate a trained encoder from these baselines. 

Thus, our contributions are as follows:
\begin{enumerate}
\item On one-step Collatz, every model fits its training set early, and it generalizes at about the same time when the training set contains enough distinct integers. The long delay appears only below a threshold number of different integers, which depends on the base (\cref{sec:data}).
\item We propose a carry measure $\eta_1$, the contraction coefficient of the carry chain over one digit, which can be computed for any base before training and predicts how many distinct integers a transformer needs to generalize in each base. Input length does not explain the ordering, and $\eta_1$ predicts which bases need fewer integers for a second Collatz map, integer division, a decoder-only transformer and three model sizes (\cref{sec:base}).
\item During the long delay, the model builds features that combine several input digits, and these features can replace training data. A new decoder trained on few integers generalizes from a frozen trained encoder and not from a random one, while with many integers the random encoder is sufficient (\cref{sec:encoder}).
\item Low-order residues are as linearly decodable from a random encoder and from the raw input as from a trained one, which adds to the known limitations of linear probes \citep{lrHewitt}. On digit tasks, high probe accuracy can come from the input digits themselves, so a probe needs a random network and the raw input as baselines (\cref{sec:decoding}).
\end{enumerate}

\section{Setup}
\label{sec:setup}

\paragraph{Task.} For a base $b$ we write an integer $n$ as its base-$b$ digits, most significant first, and the model predicts the digits of
\begin{equation}
T(n) =
\begin{cases}
n/2, & n \text{ even},\\
3n+1, & n \text{ odd}.
\end{cases}
\label{eq:collatz}
\end{equation}
Each digit is one token, and the vocabulary contains the $b$ digit tokens and three special tokens for padding, the start of the output and its end. In \cref{sec:base} we also use the map with $5n+1$ in place of $3n+1$ and integer division, $n \mapsto \lfloor n/3 \rfloor$.

\paragraph{Data.} We call one training of a model from initialization, for a given base, training-set size $N$ and seed, a run. For each run we draw a training set of $N$ integers once, uniformly with replacement from $[1, M]$, and train on that set for the whole run. In \cref{sec:data,sec:base}, $M$ is 200,000 for every run, so the input length in each base is the same at every training-set size. \Cref{apx:data} lists earlier base-8 runs that drew from other ranges. We evaluate on 5,000 integers from the same range that are excluded from the training set. At evaluation the model writes, at each position, the digit it assigns the highest probability, and a prediction counts as correct only when every output digit matches. When a training set is smaller than one batch, the data loader repeats it so that every batch contains 512 examples.

\paragraph{Model and training.} The model is an encoder-decoder transformer with six encoder layers and six decoder layers, and the encoder and the decoder each have their own learned positional embeddings. In \cref{sec:base} we also train a decoder-only transformer with twelve layers of the same width, which reads the input digits and a separator and then writes the output digits, and we train the encoder-decoder at two further model dimensions. We train with AdamW, a warmup followed by cosine decay, and gradient clipping. \Cref{apx:training} lists all settings, and \ifsubmission the code, the configuration of every run and the documents with our predictions are available at \url{https://anonymous.4open.science/r/long-delay-arithmetic-code-6210/}\else we will release the code, the configuration of every run and the training logs\fi.

\paragraph{Data threshold.} To compare bases we summarize the runs of each base and seed across training-set sizes by one number.

\defbox{Definition 1 (Data threshold)}{For a base and a seed, let $a(N)$ be the highest held-out accuracy, among evaluations every 1,000 steps up to step 30,000, of a model trained on $N$ distinct integers. The \emph{data threshold} $\Nninety$ is the number of training integers at which $a(N)$ first reaches 0.9, interpolated linearly in $\log_2 N$ between the two grid sizes on either side of 0.9.}

In a few runs the training loss rises sharply for a short time between steps 19,000 and 24,000 and accuracy recovers within a few thousand steps, and we take the highest accuracy over evaluations so that such a spike does not change the threshold.

\section{Small training sets delay generalization}
\label{sec:data}

\begin{figure}[t]
    \centering
    \includegraphics[width=0.88\linewidth]{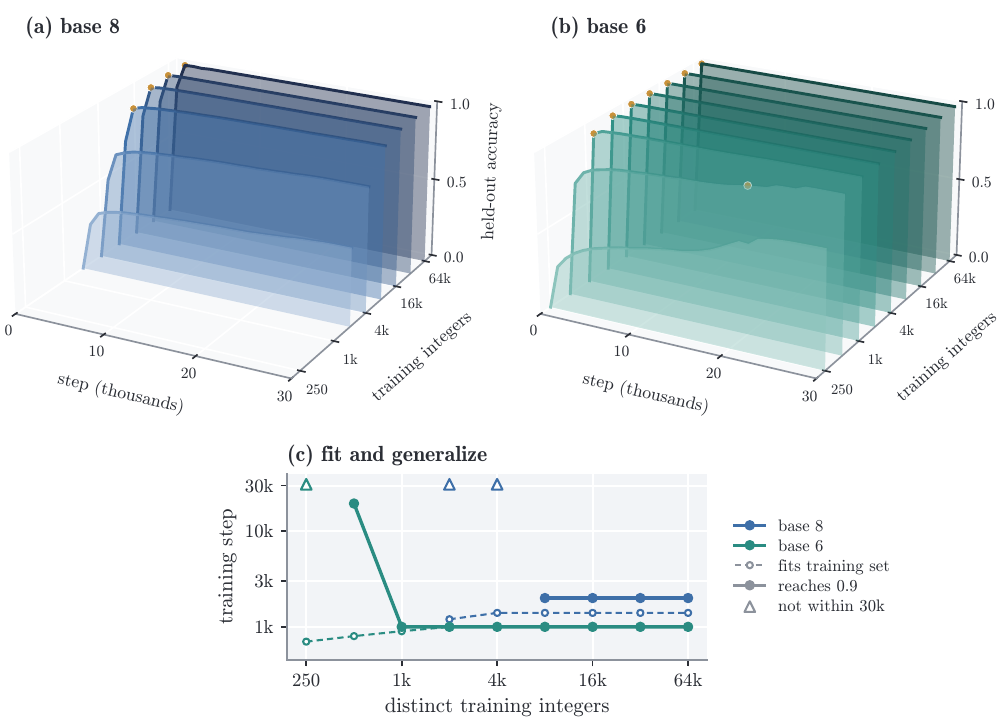}
    \caption{\textbf{Training fit and generalization in bases 8 and 6.} Held-out accuracy against training step for each training-set size in base 8 (a) and base 6 (b), averaged over 2 seeds, with a dot at the first evaluation above 0.9. Panel (c) shows, for each size, the first step with training loss below $10^{-2}$ and the first step with accuracy above 0.9. Open triangles mark sizes that stay below 0.9 for all 30,000 steps.}
    \label{fig:delay}
\end{figure}

We trained the model in bases 8 and 6 on fixed training sets that double in size, with two seeds each (\cref{fig:delay}). In both bases every run fits its training set within the first 1,500 steps, and the runs differ in how much they have generalized by then (\cref{fig:delay}c). With enough different integers the model reaches 90\% accuracy at about the time it fits. Below that size, accuracy rises slowly over tens of thousands of steps or stays below 0.9 for the whole run. The boundary lies between four and eight thousand integers in base 8 and near five hundred in base 6. Earlier base-8 runs that drew from a smaller range show the same dependence on training-set size (\cref{tab:data_regime} in \cref{apx:data}).

This is consistent with the critical dataset size that \citet{lr05} describe for modular arithmetic, and in this task the long delay is a property of the training set. The two panels also show that base 6 needs far fewer integers than base 8, which in \cref{sec:base} we examine across six bases.

\section{Bases with zero carry measure have lower data thresholds}
\label{sec:base}

\begin{figure}[t]
    \centering
    \includegraphics[width=\linewidth]{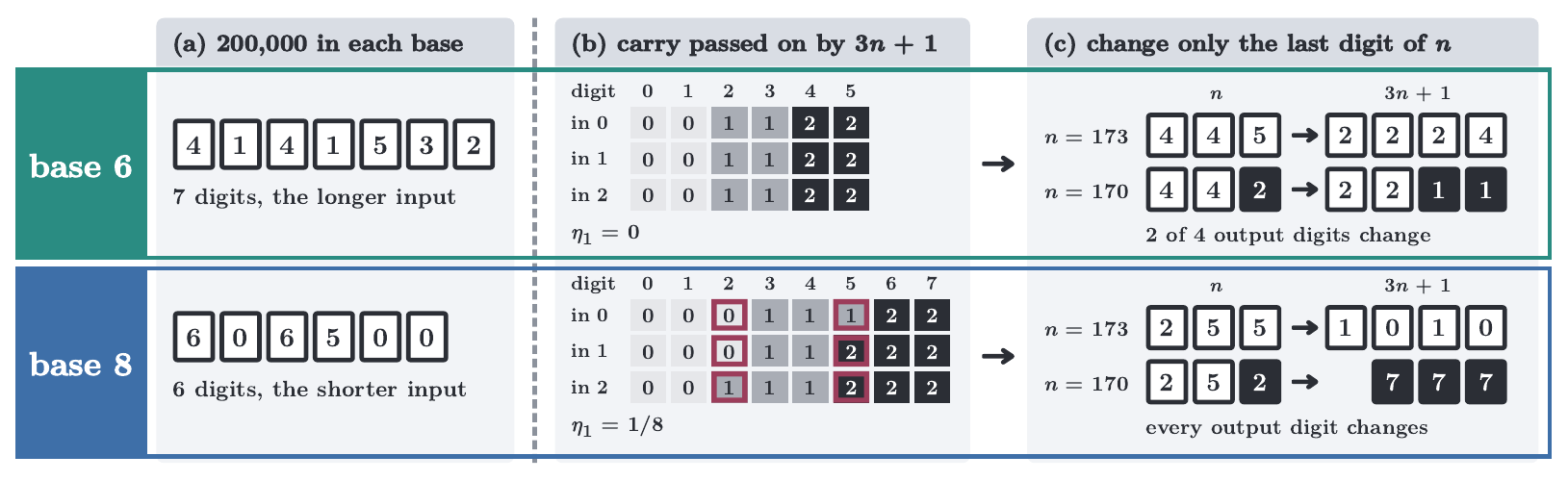}
    \caption{\textbf{An example of carries in bases 6 and 8.} (a) The input 200,000 in each base. (b) The carry each digit passes on for incoming carries 0, 1 and 2, which in base 8 depends on the incoming carry for the outlined digits. (c) $3n+1$ for $n = 173$ and $n = 170$, with filled boxes marking the digits that are different.}
    \label{fig:carry_example}
\end{figure}

\paragraph{Two predictions.} The even branch $n/2$ is computed from two adjacent digits in every even base (\cref{apx:even}), and all six bases we test are even, so the bases differ only in the odd branch $3n+1$. Concretely, to compute $3n+1$ from the least significant digit, each digit $d$ receives a carry $c$ from the digit below it, writes $(3d+c) \bmod b$ and passes the carry $\lfloor (3d+c)/b \rfloor$ to the digit above. \Cref{fig:carry_example}c shows how far a change in the last digit reaches in base 6 and in base 8, and \cref{tab:worked} in \cref{apx:carry} follows the computation digit by digit for one integer. In base 6, every digit passes on the same carry whatever carry it receives, so a change in the lowest digit can only affect the next output digit. In base 8, the digits 5 and 2 pass on a carry that depends on the carry they receive (the outlined columns of \cref{fig:carry}b), and a run of such digits lets the lowest digit decide the leading digit of the output. We summarize this difference by one number per base.

\defbox{Definition 2 (Carry measure)}{For a base $b$, let $P_c$ be the distribution of the carry passed on by a digit drawn uniformly from $\{0, \dots, b-1\}$ that receives carry $c$. The \emph{carry measure} $\eta_1(b) = \max_{c \neq c'} \mathrm{TV}(P_c, P_{c'})$ is the largest total-variation distance between these distributions. A digit is \emph{carry-sensitive} when the carry it passes on depends on the carry it receives.}

The carry measure is zero when 3 divides $b$, because every digit then passes on a carry that depends only on the digit, and it equals $1/b$ otherwise (\cref{apx:carry}). For bases 2, 8, 16 and 24 alone, $\eta_1$ and input length give the same order, so we added bases 6 and 12. Base 6 has longer inputs than base 8, and base 12 has inputs as long as those of base 16, while both have $\eta_1 = 0$. Before training, we wrote down three planned comparisons that separate the two explanations. They compare base 6 with base 8, base 6 with base 16, and base 12 with base 16. A comparison counts for a predictor when the difference in mean $\log_2 \Nninety$ has that predictor's sign and is larger than the difference between the two seeds of either base.

\paragraph{Design.} We trained bases 2, 6, 8, 12, 16 and 24 on training sets that double in size from 2,000 to 64,000 integers, with two seeds each. Before any run started, we wrote down which bases each explanation predicts to need fewer integers and the three comparisons between bases that would decide between them (\cref{apx:base}). The first results placed the threshold of bases 6, 12 and 24 below the smallest set, so following the rule we fixed before training we added smaller sets for those three bases before interpreting anything.

\begin{figure}[t]
    \centering
    \includegraphics[width=0.92\linewidth]{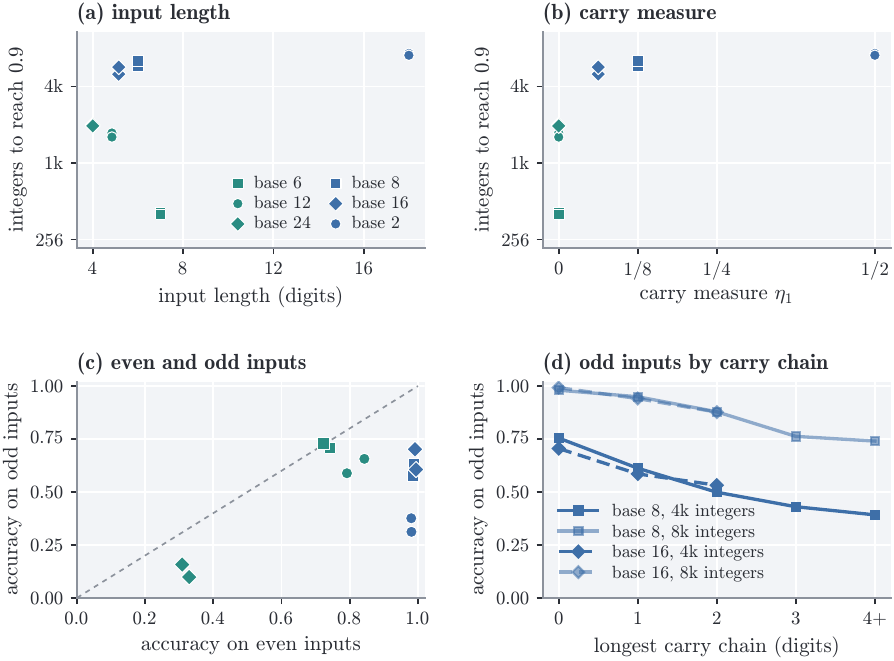}
    \caption{\textbf{Thresholds across bases.} The threshold rises with the carry measure $\eta_1$ (b) and has no consistent relation to input length (a), with one point per seed. Just below the threshold, bases 2, 8 and 16 are less accurate on odd inputs than on even ones (c, dashed line marks equal accuracy). In bases 8 and 16, accuracy on odd inputs falls as the longest run of carry-sensitive digits grows (d). Panels (c) and (d) are exploratory.}
    \label{fig:predictors}
\end{figure}

\paragraph{Results.} The bases divisible by 3 need several times fewer integers than the other three (\cref{fig:carry}c, \cref{fig:predictors}a,b and \cref{tab:base_threshold}). Moreover, all three planned comparisons favor carry propagation, each by a margin far larger than the difference between seeds, and the rank correlation of the threshold with the carry measure is high while its correlation with input length is low (\cref{tab:comparisons}). The two seeds of each base give nearly the same threshold. Since every base reaches full accuracy once the training set is large enough, the base changes the amount of data the model needs and leaves its final accuracy unchanged. This differs from the long Collatz step of \citet{lr26}, where some bases remain far from full accuracy with abundant data.

\begin{table}[t]
\caption{Bases with $\eta_1 = 0$ need fewer training integers to reach 90\% accuracy than bases with $\eta_1 > 0$. Input digits is the length of 200,000 written in the base, teal highlights mark bases where the carry measure is zero, blue highlights mark the other bases, and \cref{tab:base_full} in \cref{apx:base} gives the accuracy of every run.}
\label{tab:base_threshold}
\centering
\begin{tabular}{@{}rccrr@{}}
\toprule
Base & $\eta_1(b)$ & Input digits & $\Nninety$, seed 42 & $\Nninety$, seed 7 \\
\midrule
\zc{6} & 0 & 7 & 417 & 402 \\
\zc{12} & 0 & 5 & 1,764 & 1,638 \\
\zc{24} & 0 & 4 & 2,047 & 2,003 \\
\pc{16} & 1/16 & 5 & 5,115 & 5,812 \\
\pc{8} & 1/8 & 6 & 5,934 & 6,523 \\
\pc{2} & 1/2 & 18 & 7,451 & 7,207 \\
\bottomrule
\end{tabular}
\end{table}

\begin{table}[t]
\caption{All three comparisons planned before training count for carry propagation, and the threshold correlates strongly with the carry measure and weakly with input length. Input length predicted the opposite sign for the first two comparisons and equal values for the third. The correlations are computed across the six bases between mean $\log_2 \Nninety$ and each predictor. Base numbers are highlighted in teal where the measure is zero and in blue where it is positive, as in the figures.}
\label{tab:comparisons}
\centering
\small
\begin{tabular}{@{}lccc@{}}
\toprule
 & Difference in & Largest seed & \\
Planned comparison & mean $\log_2 \Nninety$ & difference & Counts for \\
\midrule
Base \zc{6} against base \pc{8} & $-3.92$ & 0.14 & carry propagation \\
Base \zc{6} against base \pc{16} & $-3.73$ & 0.18 & carry propagation \\
Base \zc{12} against base \pc{16} & $-1.68$ & 0.18 & carry propagation \\
\midrule
Spearman correlation with $\eta_1$ & 0.94 & & \\
Spearman correlation with input length & 0.32 & & \\
\bottomrule
\end{tabular}
\end{table}

\paragraph{Below the threshold the model errs on long carry chains.} To see where carries enter, we scored the final checkpoint of each run just below its base's threshold on every held-out integer (\cref{fig:predictors}c,d). We did not plan these analyses before training. In bases 2, 8 and 16 the model is nearly perfect on even inputs and much less accurate on odd ones, while in base 6 the two branches reach similar accuracy. Among odd inputs in bases 8 and 16, accuracy falls as the longest run of carry-sensitive digits grows, which matches the carry chain traced in \cref{tab:worked} (\cref{apx:carry}). The decline remains with twice as many training integers. In bases 12 and 24 with few integers, both branches are inaccurate, so their errors are spread over even and odd inputs alike.

\paragraph{Other operations, architectures, and model sizes.} With the multiplier 3, $\eta_1$ is zero for the bases divisible by 3 and only for them, so the experiment above cannot separate carry propagation from other properties of multiples of 3. We therefore repeated the experiment with the encoder-decoder for the map in which $5n+1$ replaces $3n+1$, and with a decoder-only transformer for $3n+1$, with comparisons planned before training (\cref{apx:b2b3}). For $5n+1$ the carry measure is zero for the bases divisible by 5, and these become the easy bases, while bases 6 and 12 become as hard as bases 8 and 16. Consequently, the easy bases follow the carry structure of the map being learned. The decoder-only transformer orders the bases as the encoder-decoder does and needs somewhat more integers. In integer division by 3, the remainder passed from digit to digit depends only on the current digit when 3 divides the base, and bases 6 and 12 again need several times fewer integers than the other bases. At a quarter and four times the parameters of the main model, base 6 still needs far fewer integers than base 8, and the larger model needs fewer integers for both bases (\cref{apx:b4b5}).

\paragraph{Limits of the base comparison.} Among the three bases with $\eta_1 = 0$ under $3n+1$, base 6 needs several times fewer integers than bases 12 and 24, and under $5n+1$ base 10 needs fewer than base 20. The carry measure gives these bases the same value, and input length orders them the opposite way. In division, only whether the measure is zero predicts the thresholds, and among the bases where it is positive the order runs against its value. After seeing the first results, we noticed that this order follows the number of digit symbols in each base, which is the number of input patterns a small training set has to cover. We did not state this prediction before training, so we report it as an observation for later work.

\section{Features learned during the delay can replace training data}
\label{sec:encoder}

\begin{figure}[t]
    \centering
    \includegraphics[width=\linewidth]{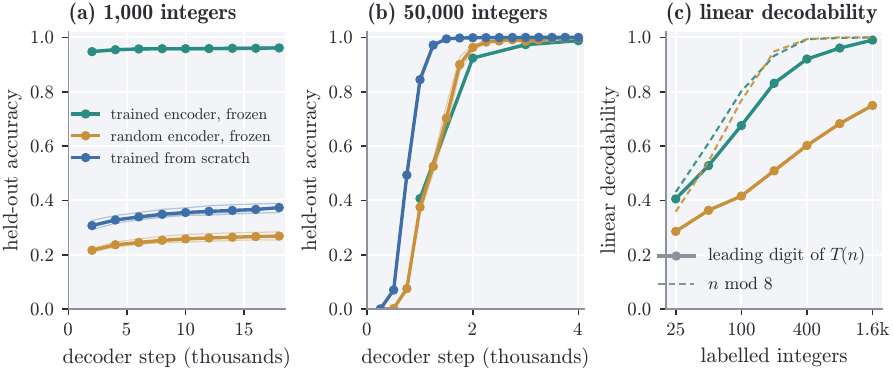}
    \caption{\textbf{Reusing a frozen encoder.} With 1,000 training integers, a new decoder reaches 96\% accuracy on the trained encoder and 27\% on a random encoder (a). Panel (b) repeats this with 50,000 training integers from $[1, 200{,}000]$, whose inputs have one digit more than the integers the encoder was trained on. A linear classifier decodes the leading digit of $T(n)$ (solid) from fewer labelled integers on the trained encoder than on the random one, shown alongside $n \bmod 8$ (dashed) (c). Thick lines are means and thin lines are individual seeds.}
    \label{fig:encoder_reuse}
\end{figure}

We took the encoder of a base-8 model that had trained to convergence on a small set of integers, froze it together with its positional embeddings, and trained a new decoder on top of it. We compared this condition with a frozen randomly initialized encoder and with training the whole model from scratch, using three data seeds for each (\cref{fig:encoder_reuse} and \cref{apx:encoder}). On the same small training set, the new decoder generalizes within a few thousand steps on the trained encoder. On the random encoder it stays below the model trained from scratch, so freezing a random feature map does not help the decoder generalize, and the gain comes from what the encoder learned.

With a large training set, the random encoder and the trained encoder lead to the same final accuracy, and training from scratch reaches it first. A decoder trained on few examples therefore relies on features the encoder learned during its own long training, while a decoder trained on many examples learns the task on top of a fixed random encoder. To see why the trained encoder helps with few examples, we fit linear classifiers to the frozen encoder states with a growing number of labelled integers (\cref{fig:encoder_reuse}c). Both encoders let a classifier read $n \bmod 8$ from about the same number of examples, since that residue is the last input digit. For the leading digit of $T(n)$, which depends on several digits, the trained encoder makes this digit decodable from a few hundred labelled integers, while its decodability from the random encoder stays lower at every number of labels we tried. A decoder trained on few integers can therefore read output features from the trained encoder that it cannot build from the random one. All of these runs use one learning rate. A decoder that trains poorly on a random encoder at this rate could train better at another, so the comparison between the two encoders holds for this learning rate.

\section{Low-order residues are decodable from the input digits without training}
\label{sec:decoding}

Linear decodability is often read as evidence that a model encodes the structure a task needs \citep{belinkov}. We fit logistic-regression classifiers to the final encoder states of a base-8 model, averaged over input positions, and compared them with the same classifiers fit to two baselines (\cref{tab:decoding}). The first baseline represents the input digits themselves, with one indicator per digit value and position, and the second is a frozen randomly initialized encoder with the same architecture. All classifiers use integers outside the model's training set, and a classifier trained on random labels scores at chance on every representation.

\begin{table}[t]
\caption{Low-order residues are decoded perfectly from the raw input and from a random encoder, and only targets that combine digit positions separate the trained encoder from these baselines. Each entry is the mean over 3 classifier seeds, every standard deviation is at most 0.016, and the model generalizes near step 44,000.}
\label{tab:decoding}
\centering
\begin{tabular}{@{}lcccc@{}}
\toprule
Representation & $n \bmod 16$ & Carry into digit 1 & Leading digit of $T(n)$ & Total carry depth \\
\midrule
One-hot input digits & 1.000 & 1.000 & 0.607 & 0.863 \\
Random encoder & 1.000 & 1.000 & 0.852 & 0.904 \\
Trained encoder, step 2,000 & 1.000 & 1.000 & 0.939 & 0.943 \\
Trained encoder, step 160,000 & 1.000 & 1.000 & 0.995 & 0.992 \\
\bottomrule
\end{tabular}
\end{table}

In base 8, the residues modulo 2, 4 and 8 are functions of the last input digit, and the residue modulo 16 adds one bit of the second-to-last digit. Any representation that keeps the position of each digit therefore supports these targets, and the trained encoder adds nothing to them at any checkpoint. Among targets that combine several digit positions, the leading digit of $T(n)$ is more decodable from the trained encoder than from the random one early in training, and its decodability keeps rising long after the model generalizes. Pretrained language models show the same pattern on this task, since the relevant residues are decodable from a randomly initialized Pythia-160M well above chance (\cref{apx:pythia}). Decodability results on digit tasks therefore need a random-network baseline and an input-only baseline before they can count as evidence of learning.

\begin{figure}[t]
    \centering
    \includegraphics[width=0.7\linewidth]{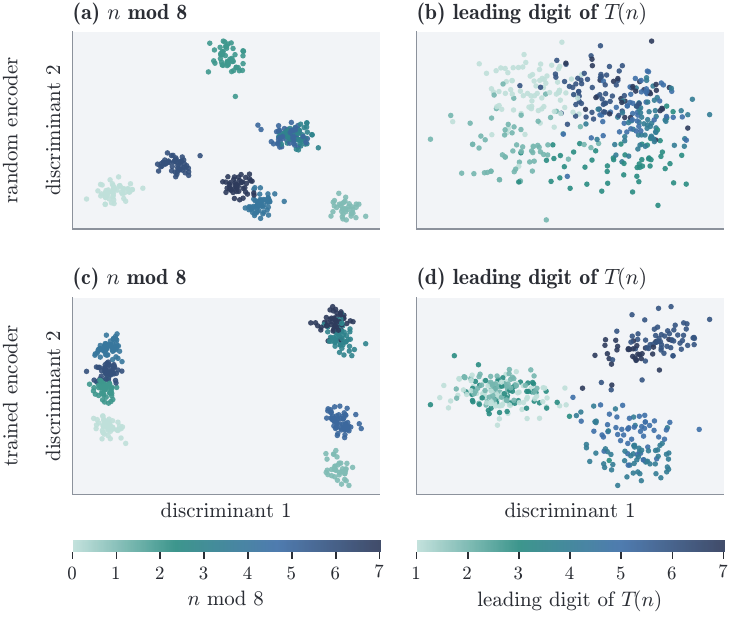}
    \caption{\textbf{Residues modulo 8 form separate clusters in the states of a random encoder as well as a trained one, and the leading digit of $T(n)$ separates only in the trained encoder.} Each panel projects the mean-pooled final encoder states of 400 held-out base-8 integers onto the two linear discriminant directions for one target, fit on 1,600 other integers. The top row uses a randomly initialized encoder and the bottom row the trained encoder at step 160,000.}
    \label{fig:decoding}
\end{figure}

\Cref{fig:decoding} shows the same contrast in the encoder states. Projected onto the two directions along which a linear method separates the values of each target best, the residues modulo 8 form distinct clusters for both encoders, while the leading digit of $T(n)$ separates only for the trained encoder.

\section{Related work}
\label{sec:related}

\paragraph{Grokking and training-set size.} Beyond the work discussed in the introduction \citep{lr01, lr05, lrZhu}, \citet{lrOG} relate the delay to weight norm across data sizes.

\paragraph{Number encoding in arithmetic transformers.} \citet{lr10} showed that the surface form of numbers decides whether sequence models learn addition, and \citet{lr11} showed that reversing the output and adding intermediate steps lower the number of examples addition needs. \citet{lr25} found that a transformer learns greatest common divisors that are products of divisors of the base, and \citet{lr26} found base-dependent accuracies for the long Collatz step with abundant data. \citet{lrDawes} showed that the structure of alternative carry functions in small bases predicts how well a recurrent network learns addition from a fixed training set. 

\paragraph{Locality and learnability.} \citet{lrAbbe} define the globality degree of a target as the number of input tokens a model must combine before its output correlates with the target, and they show that targets with high globality are hard for transformers to learn efficiently. The contraction coefficient $\eta_1$ is one concrete measure of locality for digit arithmetic, since it bounds how far a carry can influence later digits.

\paragraph{Linear decodability.} Probe accuracy is known to depend on the probe's own capacity \citep{lrHewitt} and to be high even on randomly initialized encoders \citep{zhangbowman}, and decodable properties need not be used by the model \citep{elazar}.

\section{Limitations and discussion}
\label{sec:limitations}

All results come from three digit-level operations, two transformer architectures and three model sizes, and every size uses the same learning rate and schedule. The order among bases that share a carry measure of zero is unexplained by the carry measure. Our threshold uses the best accuracy within 30,000 steps, so a base whose model would reach 90\% much later appears to need more integers. The encoder comparisons in \cref{sec:encoder} use one learning rate, and the trained encoder in the large-data condition saw shorter inputs than its decoder. The comparison of encoder checkpoints in \cref{apx:encoder} uses one seed.

We suggest that studies of grokking on arithmetic tasks report the number of distinct training examples together with the numeral base, since both change the length of the delay before generalization. We also suggest that studies using linear decodability as evidence of learning include a random network and the raw input as baselines.

\ifsubmission\else
\subsubsection*{Acknowledgments}
We thank Fran\c{c}ois Charton, whose work inspired this paper and who guided the early exploration of hypotheses. This work was supported by the Masason Foundation through its computational resources and research community.
\fi

\bibliography{main}
\bibliographystyle{tmlr}

\appendix
\crefalias{section}{appendix}
\crefalias{subsection}{appendix}

\FloatBarrier
\section{Training details}
\label{apx:training}

\begin{table}[h]
\caption{Every run in the main text uses these settings.}
\label{tab:training}
\centering
\small
\begin{tabular}{@{}lp{0.53\linewidth}@{}}
\toprule
Setting & Value \\
\midrule
Encoder and decoder layers & 6 and 6 \\
Decoder-only model & 12 layers of the same width, input digits, separator, output digits \\
Other widths (\cref{apx:b4b5}) & model dimension 128 and 512, feed-forward 512 and 2,048 \\
Model dimension, heads, feed-forward dimension & 256, 8, 1,024 \\
Positional embeddings & learned, one table for the encoder and one for the decoder \\
Optimizer & AdamW, $\beta = (0.9, 0.999)$, $\epsilon = 10^{-9}$, weight decay $10^{-2}$ \\
Learning rate & $10^{-4}$, 4,000 warmup steps, cosine decay over the run \\
Batch size & 512 \\
Gradient clipping & global norm 5 \\
Evaluation & 5,000 held-out integers, greedy decoding, exact match \\
Hardware & one NVIDIA V100 (16 GB) per run, FP32 \\
Software & Python 3.10, PyTorch 2.10 \\
\bottomrule
\end{tabular}

\end{table}

The earlier base-8 runs in \cref{apx:data} with 2,000, 3,000 and 5,000 integers use a cosine schedule over 60,000 steps, and those with 1,000 and 50,000 integers use a schedule over 200,000 steps and were stopped at steps 58,000 and 23,200. Runs in \cref{sec:data,sec:base} last 30,000 steps with an evaluation every 1,000 steps, and runs in \cref{sec:encoder} last at least 20,000 steps with 1,000 training integers and 4,000 to 10,000 steps with 50,000. Every run in \cref{sec:base} saves a checkpoint at every evaluation and resumes from it when a run is interrupted. Four of these runs were interrupted and resumed, and their logs mark the step at which they resumed.

\FloatBarrier
\section{Training-set size}
\label{apx:data}

\begin{table}[h]
\caption{Accuracy rises and the delay shortens as the number of distinct training integers grows. Each row is one seed in base 8. The 50,000-integer run ended at step 23,200, and three further runs with 1,000 integers from $[1, 10{,}000]$ reach between 0.345 and 0.375 at step 10,000.}
\label{tab:data_regime}
\centering
\begin{tabular}{@{}rrccc@{}}
\toprule
Training integers & Range & Accuracy at 10k & Accuracy at 40k & Steps to 0.9 \\
\midrule
1,000 & $[1, 10{,}000]$ & 0.349 & 0.364 & not reached by 58k \\
2,000 & $[1, 10{,}000]$ & 0.690 & 0.931 & 30,000 \\
3,000 & $[1, 10{,}000]$ & 0.837 & 0.948 & 24,000 \\
5,000 & $[1, 10{,}000]$ & 0.935 & 0.975 & 4,000 \\
1,000 & $[1, 200{,}000]$ & 0.146 & 0.439 & not reached by 60k \\
50,000 & $[1, 200{,}000]$ & 0.9996 & & 2,000 \\
\bottomrule
\end{tabular}

\end{table}

\FloatBarrier
\section{Numeral-base experiment}
\label{apx:base}

\begin{table}[h]
\caption{Each entry is the best held-out accuracy within 30,000 steps for one run of the experiment. Columns give the number of distinct training integers from $[1, 200{,}000]$. Training sets below 2,000 integers were added for bases 6, 12 and 24 after the first results, following the rule we fixed before training. Teal highlights mark bases where the measure of the map is zero and blue highlights mark the other bases.}
\label{tab:base_full}
\centering
\small
\begin{tabular}{@{}rrccccccccc@{}}
\toprule
Base & Seed & 250 & 500 & 1k & 2k & 4k & 8k & 16k & 32k & 64k \\
\midrule
\zc{6} & 42 & 0.728 & 0.961 & 0.996 & 0.999 & 1.000 & 1.000 & 1.000 & 1.000 & 1.000 \\
\zc{6} & 7 & 0.732 & 0.976 & 0.994 & 0.998 & 1.000 & 1.000 & 1.000 & 1.000 & 1.000 \\
\zc{12} & 42 & 0.033 & 0.169 & 0.688 & 0.947 & 0.995 & 1.000 & 1.000 & 1.000 & 1.000 \\
\zc{12} & 7 & 0.041 & 0.248 & 0.742 & 0.964 & 0.995 & 1.000 & 1.000 & 1.000 & 1.000 \\
\zc{24} & 42 & 0.016 & 0.066 & 0.233 & 0.897 & 0.998 & 1.000 & 1.000 & 1.000 & 1.000 \\
\zc{24} & 7 & 0.022 & 0.097 & 0.211 & 0.900 & 0.993 & 1.000 & 1.000 & 1.000 & 1.000 \\
\pc{16} & 42 &  &  &  & 0.471 & 0.853 & 0.986 & 1.000 & 1.000 & 1.000 \\
\pc{16} & 7 &  &  &  & 0.423 & 0.802 & 0.984 & 0.999 & 1.000 & 1.000 \\
\pc{8} & 42 &  &  &  & 0.493 & 0.802 & 0.974 & 0.997 & 1.000 & 1.000 \\
\pc{8} & 7 &  &  &  & 0.490 & 0.772 & 0.954 & 0.996 & 1.000 & 1.000 \\
\pc{2} & 42 &  &  &  & 0.482 & 0.636 & 0.930 & 0.988 & 0.999 & 0.999 \\
\pc{2} & 7 &  &  &  & 0.497 & 0.677 & 0.940 & 0.995 & 1.000 & 1.000 \\
\bottomrule
\end{tabular}

\end{table}

We will release, together with the code, the document in which we wrote down the predictions before any run started. Training sets of 250 and 500 integers are smaller than one batch, so the data loader repeats them to fill batches of 512, and each batch then draws from the same 250 or 500 integers.

\FloatBarrier
\subsection{The carry measure}
\label{apx:carry}

To compute $3n+1$ from the least significant digit, the model processes digit $d$ with an incoming carry $c \in \{0, 1, 2\}$, writes $(3d + c) \bmod b$ and passes the carry $\lfloor (3d + c)/b \rfloor$ to the next digit. For a window of one digit drawn uniformly from $\{0, \dots, b-1\}$, let $P_c$ be the distribution of the outgoing carry given the incoming carry $c$. We define
\begin{equation}
\eta_1(b) = \max_{c \neq c'} \mathrm{TV}(P_c, P_{c'}),
\end{equation}
which is the Dobrushin contraction coefficient of the carry chain over one digit. When $b = 3m$, the value $3d + c$ lies between $3d$ and $3d + 2$, and no multiple of $3m$ falls in that interval except possibly $3d$ itself, so the outgoing carry equals $\lfloor d/m \rfloor$ for every $c$ and $\eta_1(b) = 0$. For the bases in this paper that 3 does not divide, the computed value is $1/b$. The code we will release computes $\eta_r$ for any window $r$. \Cref{tab:worked} follows this computation for $n = 173$ in bases 6 and 8.

\begin{table}[h]
\caption{In base 6 each digit of $n = 173$ passes on the same carry whatever carry it receives, and in base 8 the digits 5 and 2 pass on a carry that depends on the carry they receive. Each row is one digit position of $n$, from the least significant, during the computation of $3n+1 = 520$. The last column of each base lists the carry the digit would pass on for incoming carries 0, 1 and 2, and bold marks the digits whose outgoing carry depends on the incoming one.}
\label{tab:worked}
\centering
\small
\setlength{\tabcolsep}{4.5pt}
\begin{tabular}{@{}c cccccc @{\hspace{14pt}} cccccc@{}}
\toprule
 & \multicolumn{6}{c}{\zc{Base 6}} & \multicolumn{6}{c}{\pc{Base 8}} \\
\cmidrule(r{8pt}){2-7} \cmidrule{8-13}
Position & $d$ & $c$ & $3d+c$ & Written & Out & Out for $c = 0, 1, 2$ & $d$ & $c$ & $3d+c$ & Written & Out & Out for $c = 0, 1, 2$ \\
\midrule
0 & 5 & 1 & 16 & 4 & 2 & 2, 2, 2 & 5 & 1 & 16 & 0 & 2 & \textbf{1, 2, 2} \\
1 & 4 & 2 & 14 & 2 & 2 & 2, 2, 2 & 5 & 2 & 17 & 1 & 2 & \textbf{1, 2, 2} \\
2 & 4 & 2 & 14 & 2 & 2 & 2, 2, 2 & 2 & 2 & 8 & 0 & 1 & \textbf{0, 0, 1} \\
3 & & 2 & & 2 & & & & 1 & & 1 & & \\
\bottomrule
\end{tabular}
\end{table}

\FloatBarrier
\subsection{The even branch}
\label{apx:even}

Write $n = \sum_i d_i b^i$ with digits $d_i \in \{0, \dots, b-1\}$, and let $e_i$ be the $i$-th digit of $n/2$. For even $b$,
\begin{equation}
e_i = \left\lfloor \frac{d_i}{2} \right\rfloor + (d_{i+1} \bmod 2) \cdot \frac{b}{2},
\end{equation}
because the parity of $d_{i+1}$ contributes $b/2$ to position $i$ and $\lfloor d_i/2 \rfloor \le b/2 - 1$ prevents any further carry. Each output digit of the even branch is therefore a function of two adjacent input digits in every even base.

\FloatBarrier
\section{Encoder reuse}
\label{apx:encoder}

The trained encoder in \cref{sec:encoder} comes from a base-8 model trained for 500,000 steps on 1,000 integers from $[1, 10{,}000]$. In that model the encoder and the decoder shared one positional table, which we copied into the encoder's own table. We froze the encoder's layers, its token embeddings and its positional embeddings, and we evaluate only on integers outside that model's training set. \Cref{tab:encoder_small,tab:encoder_large} give the values behind \cref{fig:encoder_reuse}.

\begin{table}[h]
\caption{With 1,000 training integers the frozen trained encoder leads to more than three times the accuracy of the frozen random encoder. Each entry is the mean over data seeds 42, 11 and 10.}
\label{tab:encoder_small}
\centering
\begin{tabular}{@{}lccc@{}}
\toprule
Encoder & Step 2,000 & Step 10,000 & Step 18,000 \\
\midrule
Trained, frozen & 0.947 & 0.958 & 0.961 \\
Random, frozen & 0.216 & 0.258 & 0.269 \\
Whole model trained from scratch & 0.307 & 0.356 & 0.373 \\
\bottomrule
\end{tabular}

\end{table}

\begin{table}[h]
\caption{With 50,000 training integers the model trained from scratch reaches 0.99 first. Each cell lists seeds 42 and 7. With an encoder frozen after 100 steps of training on $[1, 200{,}000]$ and seed 42, the fresh decoder reaches 0.99 at step 3,000 and ends at 0.998.}
\label{tab:encoder_large}
\centering
\begin{tabular}{@{}lccc@{}}
\toprule
Encoder & Accuracy at step 4,000 & Steps to 0.9 & Steps to 0.99 \\
\midrule
Trained on 1,000 integers, frozen & 0.986, 0.989 & 2,000, 2,000 & 5,000, 6,000 \\
Random, frozen & 0.997, 0.998 & 2,000, 1,750 & 2,750, 2,750 \\
Whole model trained from scratch & 1.000, 1.000 & 1,250, 1,250 & 1,500, 1,500 \\
\bottomrule
\end{tabular}

\end{table}

To see how the encoder's usefulness changes during its own training, we froze the same model's encoder at six checkpoints and trained a fresh decoder on each for 1,000 steps on 1,000 integers (\cref{tab:snapshot}). Later checkpoints lead to higher accuracy, and the increase continues after the model generalizes near step 44,000. This experiment uses one seed, and we omit the checkpoint at step 20,000 because it falls on a loss spike.

\begin{table}[h]
\caption{Fresh decoders reach higher accuracy on later encoder checkpoints. Each decoder trained for 1,000 steps, and the experiment uses one seed.}
\label{tab:snapshot}
\centering
\begin{tabular}{@{}lcccccc@{}}
\toprule
Encoder checkpoint & 2k & 10k & 50k & 100k & 200k & 500k \\
\midrule
Fresh-decoder accuracy & 0.303 & 0.328 & 0.596 & 0.821 & 0.929 & 0.935 \\
\bottomrule
\end{tabular}

\end{table}

\FloatBarrier
\section{Decoder-only transformer and the \texorpdfstring{$5n+1$}{5n+1} map}
\label{apx:b2b3}

We ran two further experiments with the protocol of \cref{sec:base} and fixed their comparisons before training. The first trains the decoder-only transformer on $3n+1$ in bases 6, 8, 12 and 16. The second trains the encoder-decoder on the map with $5n+1$ in bases 6, 8, 10, 12, 16 and 20. For $5n+1$ the carry measure is zero for bases 10 and 20 and positive for the other four, so carry propagation predicts that bases 10 and 20 need the fewest integers, while divisibility by 3 predicts that bases 6 and 12 keep the advantage they had under $3n+1$.

\begin{figure}[h]
    \centering
    \includegraphics[width=\linewidth]{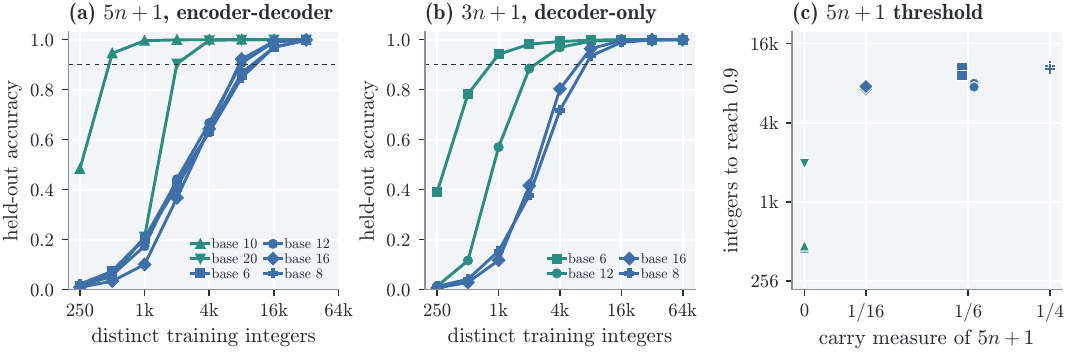}
    \caption{\textbf{Under $5n+1$ the bases divisible by 5 need the fewest training integers, and a decoder-only transformer orders the bases as the encoder-decoder does under $3n+1$.} Teal marks bases where the carry measure of the map is zero and blue marks the other bases. Panels (a) and (b) show the mean over 2 seeds of the best held-out accuracy within 30,000 steps, and the dashed line marks 0.9. Panel (c) shows the integers needed to reach 0.9 under $5n+1$ against the carry measure of $5n+1$, and each point is one seed.}
    \label{fig:b2b3}
\end{figure}

In both experiments every planned comparison counts for carry propagation (\cref{tab:b2b3_comparisons}). Under $5n+1$ the easy bases change to 10 and 20, and bases 6 and 12 need about as many integers as bases 8 and 16 (\cref{fig:b2b3}). The threshold correlates strongly with the carry measure of $5n+1$ and shows no correlation with the carry measure of $3n+1$, so the carry structure of the map being learned sets how much data a base needs. The decoder-only transformer orders bases 6, 12, 16 and 8 as the encoder-decoder does, and it needs somewhat more integers for the easy bases (\cref{tab:b2b3_threshold}). Among the bases where the carry measure is zero, base 10 needs fewer integers than base 20, in the same direction as the digit-symbol observation of \cref{sec:base}.

\begin{table}[h]
\caption{Thresholds of the two further experiments. Each entry is the number of training integers at which the best held-out accuracy first reaches 0.9. Teal highlights mark bases where the carry measure of the row's map is zero and blue highlights mark the other bases.}
\label{tab:b2b3_threshold}
\centering
\small
\begin{tabular}{@{}lcccccc@{}}
\toprule
 & \multicolumn{6}{c}{Base} \\
\cmidrule(l){2-7}
 & 6 & 8 & 10 & 12 & 16 & 20 \\
\midrule
$5n+1$, encoder-decoder, seed 42 & \pc{10,623} & \pc{11,023} & \zc{458} & \pc{8,165} & \pc{7,393} & \zc{1,989} \\
$5n+1$, encoder-decoder, seed 7 & \pc{9,356} & \pc{10,511} & \zc{474} & \pc{7,648} & \pc{7,725} & \zc{1,996} \\
$3n+1$, decoder-only, seed 42 & \zc{859} & \pc{7,262} &  & \zc{2,527} & \pc{6,192} &  \\
$3n+1$, decoder-only, seed 7 & \zc{819} & \pc{7,079} &  & \zc{1,993} & \pc{5,968} &  \\
\bottomrule
\end{tabular}
\end{table}

\begin{table}[h]
\caption{Every comparison planned for the two further experiments counts for carry propagation. The correlations are Spearman correlations across the bases of each experiment between mean $\log_2 \Nninety$ and each predictor. Base numbers are highlighted in teal where the measure is zero and in blue where it is positive, as in the figures.}
\label{tab:b2b3_comparisons}
\centering
\small
\begin{tabular}{@{}llccc@{}}
\toprule
 & & Difference in & Largest seed & \\
Experiment & Planned comparison & mean $\log_2 \Nninety$ & difference & Counts for \\
\midrule
$5n+1$, & Base \zc{10} against base \pc{6} & $-4.42$ & 0.18 & carry propagation \\
encoder-decoder & Base \zc{20} against base \pc{12} & $-1.99$ & 0.09 & carry propagation \\
 & Base \zc{10} against base \pc{8} & $-4.53$ & 0.07 & carry propagation \\
 & Correlation with carry measure of $5n+1$ & 0.97 & & \\
 & Correlation with carry measure of $3n+1$ & $-0.06$ & & \\
 & Correlation with input length & 0.37 & & \\
\midrule
$3n+1$, & Base \zc{6} against base \pc{8} & $-3.10$ & 0.07 & carry propagation \\
decoder-only & Base \zc{6} against base \pc{16} & $-2.86$ & 0.07 & carry propagation \\
 & Base \zc{12} against base \pc{16} & $-1.44$ & 0.34 & carry propagation \\
 & Correlation with carry measure & 0.95 & & \\
 & Correlation with input length & $-0.32$ & & \\
\bottomrule
\end{tabular}
\end{table}

\begin{table}[h]
\caption{Each entry is the best held-out accuracy within 30,000 steps for one run of the $5n+1$ experiment. Columns give the number of distinct training integers from $[1, 200{,}000]$. Teal highlights mark bases where the measure of the map is zero and blue highlights mark the other bases.}
\label{tab:b3_full}
\centering
\small
\begin{tabular}{@{}rrcccccccc@{}}
\toprule
Base & Seed & 250 & 500 & 1k & 2k & 4k & 8k & 16k & 32k \\
\midrule
\zc{10} & 42 & 0.541 & 0.952 & 0.995 & 1.000 & 1.000 & 1.000 & 1.000 & 1.000 \\
\zc{10} & 7 & 0.424 & 0.939 & 0.996 & 0.999 & 1.000 & 1.000 & 1.000 & 1.000 \\
\zc{20} & 42 & 0.018 & 0.059 & 0.228 & 0.906 & 0.998 & 1.000 & 1.000 & 1.000 \\
\zc{20} & 7 & 0.010 & 0.047 & 0.195 & 0.902 & 0.996 & 1.000 & 1.000 & 1.000 \\
\pc{6} & 42 & 0.018 & 0.088 & 0.212 & 0.424 & 0.638 & 0.855 & 0.965 & 0.996 \\
\pc{6} & 7 & 0.010 & 0.061 & 0.193 & 0.398 & 0.633 & 0.878 & 0.974 & 0.999 \\
\pc{8} & 42 & 0.026 & 0.067 & 0.191 & 0.422 & 0.627 & 0.843 & 0.966 & 0.998 \\
\pc{8} & 7 & 0.018 & 0.079 & 0.219 & 0.445 & 0.631 & 0.852 & 0.974 & 0.998 \\
\pc{12} & 42 & 0.014 & 0.052 & 0.155 & 0.423 & 0.657 & 0.897 & 0.992 & 0.999 \\
\pc{12} & 7 & 0.019 & 0.064 & 0.193 & 0.460 & 0.675 & 0.916 & 0.990 & 0.998 \\
\pc{16} & 42 & 0.009 & 0.030 & 0.069 & 0.394 & 0.676 & 0.929 & 0.992 & 0.999 \\
\pc{16} & 7 & 0.009 & 0.037 & 0.131 & 0.339 & 0.611 & 0.915 & 0.994 & 1.000 \\
\bottomrule
\end{tabular}

\end{table}

\begin{table}[h]
\caption{Each entry is the best held-out accuracy within 30,000 steps for one run of the decoder-only experiment. Columns give the number of distinct training integers from $[1, 200{,}000]$. Teal highlights mark bases where the measure of the map is zero and blue highlights mark the other bases.}
\label{tab:b2_full}
\centering
\small
\begin{tabular}{@{}rrccccccccc@{}}
\toprule
Base & Seed & 250 & 500 & 1k & 2k & 4k & 8k & 16k & 32k & 64k \\
\midrule
\zc{6} & 42 & 0.460 & 0.790 & 0.931 & 0.982 & 0.992 & 0.999 & 1.000 & 1.000 & 1.000 \\
\zc{6} & 7 & 0.324 & 0.774 & 0.951 & 0.981 & 0.994 & 0.998 & 1.000 & 1.000 & 1.000 \\
\zc{12} & 42 & 0.015 & 0.100 & 0.533 & 0.867 & 0.965 & 0.995 & 0.999 & 0.999 & 1.000 \\
\zc{12} & 7 & 0.013 & 0.132 & 0.607 & 0.901 & 0.974 & 0.993 & 0.998 & 0.999 & 1.000 \\
\pc{8} & 42 & 0.013 & 0.037 & 0.144 & 0.392 & 0.714 & 0.930 & 0.992 & 0.999 & 1.000 \\
\pc{8} & 7 & 0.012 & 0.046 & 0.164 & 0.362 & 0.722 & 0.938 & 0.986 & 0.999 & 1.000 \\
\pc{16} & 42 & 0.007 & 0.026 & 0.106 & 0.406 & 0.801 & 0.958 & 0.996 & 1.000 & 1.000 \\
\pc{16} & 7 & 0.006 & 0.032 & 0.128 & 0.425 & 0.804 & 0.970 & 0.996 & 1.000 & 1.000 \\
\bottomrule
\end{tabular}

\end{table}

\FloatBarrier
\section{Integer division and model size}
\label{apx:b4b5}

We ran two more experiments with the protocol of \cref{sec:base} and fixed their comparisons before training. The first trains the encoder-decoder on integer division, $n \mapsto \lfloor n/3 \rfloor$, in bases 6, 8, 10, 12, 16 and 20. Long division passes a remainder from the most significant digit to the least significant one, so it travels in the direction the model writes its output, while the carry of $3n+1$ travels the other way. When 3 divides the base, the remainder passed on depends only on the current digit, so the measure of \cref{apx:carry}, applied to the remainder, is zero for bases 6 and 12 and positive for the other four. The second experiment trains the encoder-decoder on $3n+1$ in bases 6 and 8 at model dimensions 128 and 512, which have about a quarter and four times the parameters of the main model.

\begin{figure}[h]
    \centering
    \includegraphics[width=0.9\linewidth]{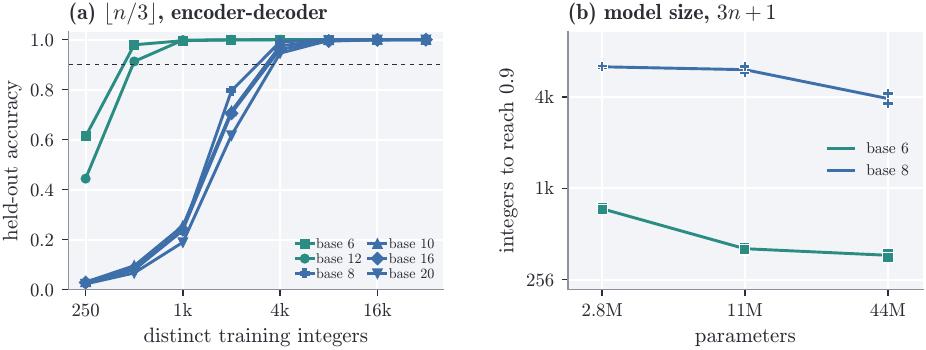}
    \caption{\textbf{In division by 3 the bases divisible by 3 need about 7 times fewer training integers than the other bases, and base 6 needs about 9 to 15 times fewer integers than base 8 at every model size.} Panel (a) shows the mean over 2 seeds of the best held-out accuracy within 30,000 steps for $\lfloor n/3 \rfloor$. Teal marks bases where the remainder passed between digits depends only on the current digit, and the dashed line marks 0.9. Panel (b) shows the integers needed to reach 0.9 for $3n+1$ against the number of parameters, with one point per seed and a line through the geometric means.}
    \label{fig:b4b5}
\end{figure}

In division, every planned comparison favours bases 6 and 12 (\cref{tab:b4b5_comparisons}), and these two bases need several times fewer integers than the other four (\cref{fig:b4b5}a). We had expected a flat result, since the remainder travels in the direction the output is written, so the base effect does not depend on the direction of propagation. Among the four bases where the measure is positive, the thresholds lie close together and rise with the base, the opposite of the measure's value. In division the measure therefore separates two groups of bases and does not order the bases within the second group.

At both further model sizes, base 6 needs far fewer integers than base 8 (\cref{fig:b4b5}b). The threshold falls as the model grows for both bases, so on this task a larger model needs fewer distinct integers, and the gap between the two bases stays large at every size. All three sizes use the same learning rate and schedule.

\begin{table}[h]
\caption{Thresholds of the division and model-size experiments. Each entry is the number of training integers at which the best held-out accuracy first reaches 0.9, for seeds 42 and 7. Teal highlights mark bases where the carry or remainder measure of the row's map is zero and blue highlights mark the other bases.}
\label{tab:b4b5_threshold}
\centering
\small
\begin{tabular}{@{}lcccccc@{}}
\toprule
 & \multicolumn{6}{c}{Base} \\
\cmidrule(l){2-7}
 & 6 & 8 & 10 & 12 & 16 & 20 \\
\midrule
$\lfloor n/3 \rfloor$, seed 42 & \zc{434} & \pc{3,156} & \pc{3,244} & \zc{534} & \pc{3,390} & \pc{3,541} \\
$\lfloor n/3 \rfloor$, seed 7 & \zc{425} & \pc{2,408} & \pc{3,306} & \zc{472} & \pc{3,381} & \pc{3,711} \\
\midrule
$3n+1$, width 128 (2.8M), seeds 42 and 7 & \zc{762, 740} & \pc{6,433, 6,565} &  &  &  &  \\
$3n+1$, width 256 (11.1M), seeds 42 and 7 & \zc{417, 402} & \pc{5,934, 6,523} &  &  &  &  \\
$3n+1$, width 512 (44.2M), seeds 42 and 7 & \zc{381, 360} & \pc{3,711, 4,334} &  &  &  &  \\
\bottomrule
\end{tabular}
\end{table}

\begin{table}[h]
\caption{Every comparison planned for the division and model-size experiments counts for the measure. The width-256 row repeats the corresponding comparison from \cref{sec:base} for reference. Base numbers are highlighted in teal where the measure is zero and in blue where it is positive, as in the figures.}
\label{tab:b4b5_comparisons}
\centering
\small
\begin{tabular}{@{}llccc@{}}
\toprule
 & & Difference in & Largest seed & \\
Experiment & Planned comparison & mean $\log_2 \Nninety$ & difference & Counts for \\
\midrule
$\lfloor n/3 \rfloor$ & Base \zc{6} against base \pc{8} & $-2.68$ & 0.39 & remainder measure \\
 & Base \zc{12} against base \pc{16} & $-2.75$ & 0.18 & remainder measure \\
 & Base \zc{12} against base \pc{20} & $-2.85$ & 0.18 & remainder measure \\
\midrule
$3n+1$, model size & Base \zc{6} against base \pc{8}, width 128 & $-3.11$ & 0.04 & carry measure \\
 & Base \zc{6} against base \pc{8}, width 256 & $-3.92$ & 0.14 & carry measure \\
 & Base \zc{6} against base \pc{8}, width 512 & $-3.44$ & 0.22 & carry measure \\
\bottomrule
\end{tabular}
\end{table}

\begin{table}[h]
\caption{Each entry is the best held-out accuracy within 30,000 steps for one run of the division experiment. Columns give the number of distinct training integers from $[1, 200{,}000]$. Teal highlights mark bases where the measure of the map is zero and blue highlights mark the other bases.}
\label{tab:b4_full}
\centering
\small
\begin{tabular}{@{}lrcccccccc@{}}
\toprule
Base & Seed & 250 & 500 & 1k & 2k & 4k & 8k & 16k & 32k \\
\midrule
\zc{6} & 42 & 0.565 & 0.985 & 0.997 & 1.000 & 1.000 & 1.000 & 1.000 & 1.000 \\
\zc{6} & 7 & 0.660 & 0.973 & 0.995 & 0.998 & 1.000 & 1.000 & 1.000 & 1.000 \\
\zc{12} & 42 & 0.379 & 0.890 & 0.998 & 1.000 & 1.000 & 1.000 & 1.000 & 1.000 \\
\zc{12} & 7 & 0.508 & 0.936 & 0.998 & 0.999 & 1.000 & 1.000 & 1.000 & 1.000 \\
\pc{8} & 42 & 0.020 & 0.079 & 0.223 & 0.725 & 0.991 & 1.000 & 1.000 & 1.000 \\
\pc{8} & 7 & 0.031 & 0.080 & 0.250 & 0.867 & 0.991 & 1.000 & 1.000 & 1.000 \\
\pc{10} & 42 & 0.024 & 0.085 & 0.213 & 0.725 & 0.976 & 0.999 & 1.000 & 1.000 \\
\pc{10} & 7 & 0.036 & 0.106 & 0.297 & 0.703 & 0.975 & 0.998 & 1.000 & 1.000 \\
\pc{16} & 42 & 0.030 & 0.072 & 0.246 & 0.717 & 0.957 & 0.995 & 1.000 & 1.000 \\
\pc{16} & 7 & 0.027 & 0.094 & 0.234 & 0.694 & 0.966 & 0.994 & 0.999 & 1.000 \\
\pc{20} & 42 & 0.023 & 0.062 & 0.214 & 0.647 & 0.954 & 0.998 & 1.000 & 1.000 \\
\pc{20} & 7 & 0.022 & 0.071 & 0.164 & 0.588 & 0.938 & 0.996 & 1.000 & 1.000 \\
\bottomrule
\end{tabular}

\end{table}

\begin{table}[h]
\caption{Each entry is the best held-out accuracy within 30,000 steps for one run of the model-size experiment. Rows give the model dimension and the base, and columns give the number of distinct training integers from $[1, 200{,}000]$. Teal highlights mark bases where the measure of the map is zero and blue highlights mark the other bases.}
\label{tab:b5_full}
\centering
\small
\begin{tabular}{@{}lrccccccccc@{}}
\toprule
Width, base & Seed & 250 & 500 & 1k & 2k & 4k & 8k & 16k & 32k & 64k \\
\midrule
\zc{128, base 6} & 42 & 0.508 & 0.806 & 0.960 & 0.997 & 0.999 & 1.000 & 1.000 & 1.000 & 1.000 \\
\zc{128, base 6} & 7 & 0.425 & 0.811 & 0.968 & 0.996 & 0.999 & 1.000 & 1.000 & 1.000 & 1.000 \\
\pc{128, base 8} & 42 & 0.007 & 0.020 & 0.150 & 0.446 & 0.750 & 0.969 & 0.995 & 1.000 & 1.000 \\
\pc{128, base 8} & 7 & 0.004 & 0.025 & 0.172 & 0.462 & 0.760 & 0.956 & 0.996 & 1.000 & 1.000 \\
\zc{512, base 6} & 42 & 0.795 & 0.967 & 0.998 & 0.999 & 1.000 & 1.000 & 1.000 & 1.000 & 1.000 \\
\zc{512, base 6} & 7 & 0.803 & 0.988 & 0.998 & 0.999 & 1.000 & 1.000 & 1.000 & 1.000 & 1.000 \\
\pc{512, base 8} & 42 & 0.013 & 0.056 & 0.373 & 0.842 & 0.907 & 0.975 & 0.998 & 1.000 & 1.000 \\
\pc{512, base 8} & 7 & 0.009 & 0.060 & 0.429 & 0.827 & 0.892 & 0.964 & 0.997 & 1.000 & 1.000 \\
\bottomrule
\end{tabular}

\end{table}

\FloatBarrier
\section{Decodability in other architectures}
\label{apx:pythia}

\paragraph{Pythia-160M.} We fit logistic-regression classifiers to the residual stream of Pythia-160M \citep{lrPythia} at the last input position of base-8 Collatz inputs. We repeated this at nine pretraining checkpoints and at random initialization. On the randomly initialized network the classifiers recover $n \bmod 2$, $n \bmod 4$, $n \bmod 8$ and the last digit with accuracy 1.00, and $n \bmod 16$ and $T(n) \bmod 8$ with accuracy between 0.73 and 0.94. On the pretrained network they recover the latter targets with accuracy 1.00 from pretraining step 1,000 onward. Most of this decodability therefore comes from the input, and the rest forms early in pretraining.

\paragraph{Pythia-1.4B.} After full fine-tuning on base-8 Collatz, exact-match accuracy rises from 0 at step 0 to at least 0.97 by step 500. At step 0 a linear classifier recovers $T(n) \bmod 8$ and $n \bmod 16$ with accuracy 1.00. Both targets are functions of the last two input digits, and we did not fit classifiers to a randomly initialized 1.4B network.

\FloatBarrier
\end{document}